\documentclass[runningheads]{llncs}

\usepackage[hidelinks, colorlinks=true, linkcolor=blue, citecolor=blue, breaklinks=true]{hyperref}
\usepackage[T1]{fontenc}
\usepackage{graphicx}
\usepackage{multirow}
\usepackage{xcolor}
\usepackage{colortbl}
\usepackage{booktabs}
\usepackage{ulem}
\usepackage{pifont}
\usepackage{amsfonts,amssymb,amsmath}
\usepackage{dsfont}
\usepackage{float}
\usepackage{subfigure}
\usepackage{subcaption}
\usepackage{marvosym}
\usepackage[utf8]{inputenc}

\begin{document}

\title{PG-SAM: Prior-Guided SAM with Medical for Multi-organ Segmentation}
\author{
Yiheng~Zhong\textsuperscript{\dag}\inst{1,2,3}\and
Zihong~Luo\textsuperscript{\dag}\inst{1,2,3}\and
Chengzhi~Liu\inst{2,3} \and
Feilong~Tang\inst{1,4} \and
Zelin~Peng\inst{5} \and
Ming~Hu\inst{4} \and
Yingzhen~Hu\inst{2} \and
Jionglong~Su\inst{2} \and \\
Zongyuan~Ge\textsuperscript{\Letter}\inst{4} \and
Imran~Razzak\textsuperscript{\Letter}\inst{1}
}
\institute{Mohamed bin Zayed University of AI, Abu Dhabi, UAE \and Xi'an Jiaotong-Liverpool University, Suzhou, China \and University of Liverpool, Liverpool, United Kingdom \and Monash University, Melbourne, Australia \and Shanghai Jiao Tong University, Shanghai, China \\ \email{imran.razzak@mbzuai.ac.ae}} 
\maketitle


\begingroup
\renewcommand\thefootnote{}\footnotetext{\textsuperscript{\dag}Equal contribution.}
\renewcommand\thefootnote{}\footnotetext{\textsuperscript{\Letter}Corresponding author.}
\endgroup

\begin{abstract}
Segment Anything Model (SAM) demonstrates powerful zero-shot capabilities; however, its accuracy and robustness significantly decrease when applied to medical image segmentation. Existing methods address this issue through modality fusion, integrating textual and image information to provide more detailed priors. In this study, we argue that the granularity of text and the domain gap affect the accuracy of the priors. Furthermore, the discrepancy between high-level abstract semantics and pixel-level boundary details in images can introduce noise into the fusion process. To address this, we propose Prior-Guided SAM (\textbf{PG-SAM}), which employs a fine-grained modality prior aligner to leverage specialized medical knowledge for better modality alignment. The core of our method lies in efficiently addressing the domain gap with fine-grained text from a medical LLM. Meanwhile, it also enhances the priors' quality after modality alignment, ensuring more accurate segmentation. In addition, our decoder enhances the model’s expressive capabilities through multi-level feature fusion and iterative mask optimizer operations, supporting unprompted learning. We also propose a unified pipeline that effectively supplies high-quality semantic information to SAM. Extensive experiments on the Synapse dataset demonstrate that the proposed PG-SAM achieves state-of-the-art performance. Our code is released at  \url{https://github.com/logan-0623/PG-SAM}.

\keywords{SAM \and Prompt-free Multi-organ Segmentation \and LLM}
\end{abstract}

\vspace{-0.2cm}
\section{Introduction}
\vspace{-0.2cm}
\label{sec:intro}

Multi-organ segmentation is a core task in medical image analysis, aiming to accurately separate multiple organs. Segment Anything Model (SAM)~\cite{kirillov2023segment} demonstrating its broad application potential~\cite{he2023computer,hu2023skinsam,ji2023sam,li2024polyp,mazurowski2023segment,tang2024hunting,wu2025mswal}. The success of SAM relies on precise prompts. However, traditional SAM methods are time-consuming, rely on domain expertise, and are prone to human error~\cite{feng2023cheap,li2024tp,wu2023medical,zhang2023self}. To address these challenges, recent research has focused on prompt-free methods to offer simpler and more efficient segmentation solutions~\cite{chen2024sam,he2025few,xie2024masksam,zhang2023customized,tang2024neighbor}. These methods leverage prior information to aid the decoder in better segmentation.

Inspired by multimodal learning, recent methods exploit textual information to generate priors for enhancing segmentation~\cite{aleem2024test,chen2024test,huang2024seg,li2024tp}. However, these approaches overlook the fact that the semantic representations derived from visual–language models (VLMs) are largely abstract and non-pixel-level, which can introduce noise. Moreover, the granularity of text descriptions influences the quality of the priors; coarse descriptions result in poor alignment with image features, thereby undermining segmentation accuracy~\cite{li2024tp}. In this work, we propose that narrowing the gap between semantic information and pixel-level boundary information can improve segmentation and mitigate noise.

\begin{figure}[t]
    \centering
    \includegraphics[width=0.93\textwidth]{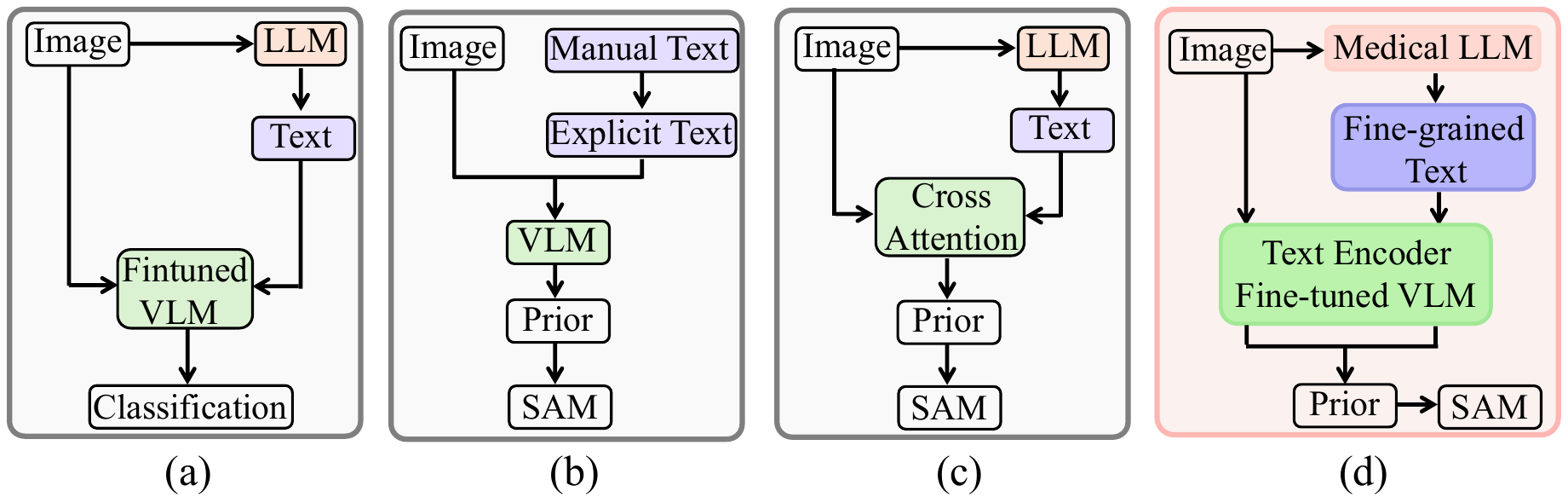}  
    \vspace{-0.3cm}
    \caption{Comparison of PG-SAM with other methods: (a) Issues with text granularity. (b) Fine-grained explicit text relies on manual verification and suffers from alignment problems. (c) Faces text granularity issues and lacks the zero-shot capabilities of VLM. (d) Our pipeline improves modality alignment through VLM fine-tuning, providing fine-grained text and the most comprehensive improves.}
    \label{fig:1}
    \vspace{-0.3cm}
\end{figure}

Text-visual alignment methods have shown potential in guiding SAM. Prompt learning-based methods~\cite{fang2024aligning} generate visual descriptions by aligning with CLIP~\cite{radford2021learning}, providing valuable text-visual information. However, these methods suffer from text granularity limitations, which compromise modality alignment. Furthermore, the absence of a dedicated segmentation stage makes capturing fine-grained image details challenging, as illustrated in Fig.~\ref{fig:1} (a). In contrast, TP-DRSeg~\cite{li2024tp} directly aligns explicit text with CLIP and generates priors to assist SAM, offering finer-grained information. However, it relies on ophthalmology texts verified by experts and faces alignment issues due to the VLM’s reliance on natural-image training, as shown in Fig.~\ref{fig:1} (b). Meanwhile, SEG-SAM~\cite{huang2024seg} employs LLMs to provide text information, leveraging cross-attention to calculate similarity and more efficiently introduce semantic information into SAM, as depicted in Fig.~\ref{fig:1} (c). However, it still faces granularity issues and does not fully exploit the zero-shot capability of VLM, potentially affecting its generalization.

To this end, we propose Prior-Guided SAM (PG-SAM), an efficient pipeline that provides domain-adapted, fine-grained priors and mitigates domain gap issues, as shown in Fig.~\ref{fig:1} (d). Specifically, we introduce a fine-grained modality prior aligner that leverages medical LLMs to merge the fluency of large-scale models with the domain-specific expertise of medical professionals, thereby excelling in complex medical scenarios and delivering more detailed, specific semantic information to CLIP. Furthermore, we fine-tune CLIP with Low-Rank Adaptation (LoRA)~\cite{hu2022lora} for the medical domain, providing more accurate semantic priors. While these priors complement the embeddings of SAM, image features of CLIP focus on abstract semantics, lacking pixel-level details, which may cause simple fusion methods to blur SAM’s embeddings and hinder the retention of fine-grained features. To address this, we design a novel decoder that enhances feature extraction through multi-level fusion, reducing detail loss from noise and promoting knowledge sharing between the CLIP and SAM. Finally, by leveraging an iterative mask optimizer, we dynamically fine-tune the mask weights for each category, enhancing feature expression and enabling better discrimination of small organ details. Extensive experiments on the Synapse dataset demonstrate that the proposed PG-SAM achieves state-of-the-art performance.

Overall, our contributions are threefold: (1) We propose a fine-grained modality prior aligner that combines high-level semantics and visual information to generate high-quality priors for all categories; (2) We introduce a novel decoder, which improves mask quality through multi-level feature fusion and a mask fine-tuner; (3) We provide a unified pipeline that simplifies the process while enriching prompt-free methods with medical knowledge. Experimental results demonstrate that our method improves multi-organ segmentation, outperforming state-of-the-art performance on the Synapse dataset~\cite{Landman2015}.

\vspace{-0.2cm}
\section{Methodology}
\vspace{-0.2cm}
\subsection{Overview}
\vspace{-0.2cm}

\begin{figure}[t]
    \centering
    \includegraphics[width=\textwidth]{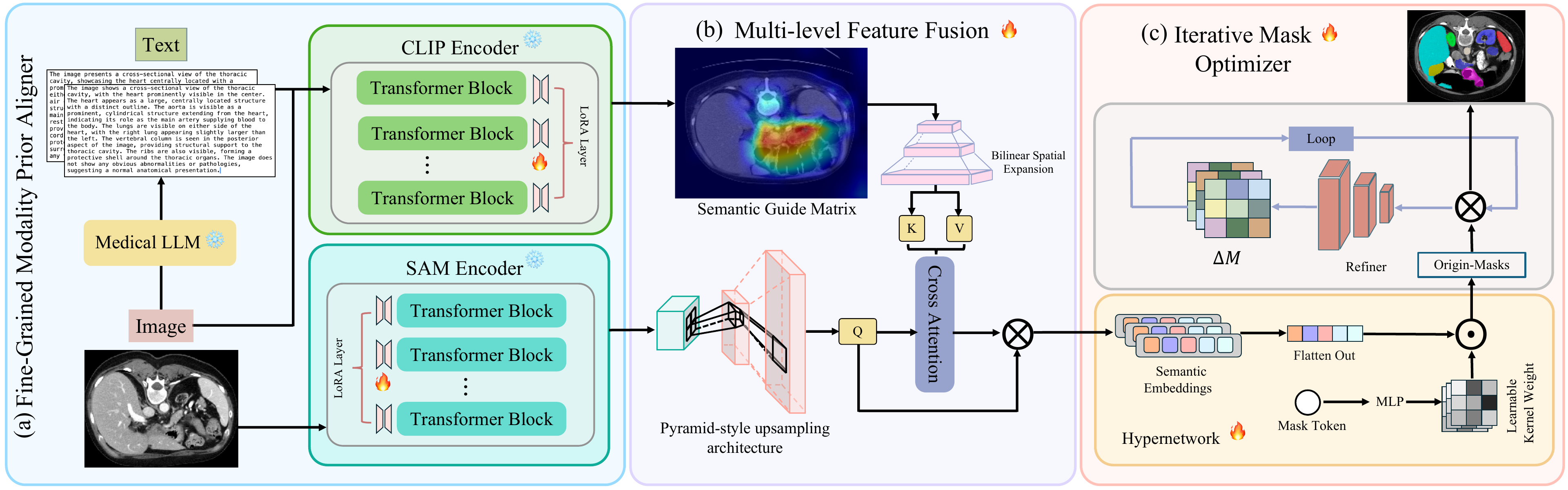} 
    \vspace{-0.5cm}
    \caption{Overview of PG-SAM. (a) Illustrates the process by which the fine-grained modality prior aligner generates the Semantic Guide Matrix $G$; (b) For multi-level feature fusion, $G$ is integrated with the feature map after multi-level sampling to preserve more detailed features; (c) It outlines the iterative mask optimizer, which dynamically learns convolution kernel parameters via a Hypernetwork and refines the final mask using a dedicated refiner.}
    \label{fig2}
    \vspace{-0.3cm}
\end{figure}

Fig.~\ref{fig2} illustrates the overview of our method, consisting of three coordinated key components: a \textit{Fine-Grained Modality Prior Aligner}, as described in Section~\ref{sec:module1}, a \textit{Multi-level Feature Fusion}, detailed in Section~\ref{sec:module2}, and a \textit{Iterative Mask Optimizer}, explained in Section~\ref{sec:module3}. PG-SAM first generates fine-grained text descriptions for each image, combining them to generate semantic priors, referred to as the Semantic Guide Matrix, to assist the decoding process. Then the Multi-level Feature Fusion module facilitate knowledge sharing between the text-guided explicit prior and multi-level visual features. Finally, in the iterative mask optimizer, candidate masks are provided for each category via mask tokens, and a mask refiner optimizes these segmentation details.

\vspace{-0.2cm}
\subsection{Fine-Grained Modality Prior Aligner} \label{sec:module1}
\vspace{-0.2cm}

\begin{figure}[t]
    \centering
    \includegraphics[width=\textwidth]{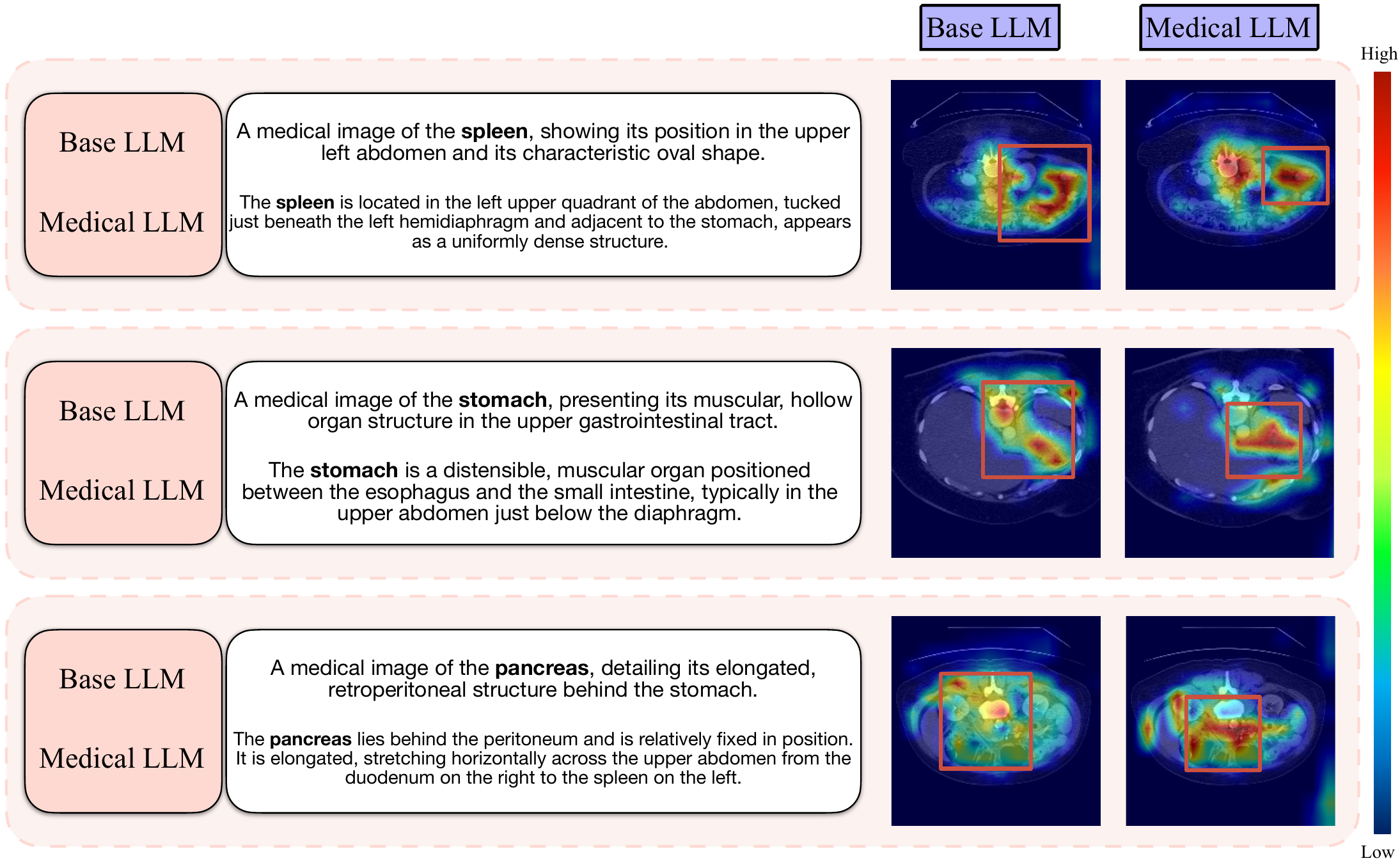} 
    \vspace{-0.5cm}
    \caption{Comparison of textual prompts and corresponding heatmaps generated by the Base LLM (left) and the Medical LLM (right) for anatomical images of the \textit{spleen}, \textit{stomach}, and \textit{pancreas}. The Medical LLM provides clinically precise descriptions, yielding more focused and detailed semantic guidance, as demonstrated by the sharper heatmap regions.}
    \label{pic2}
    \vspace{-0.5cm}
\end{figure}

The aligner employs a Medical-LLM to generate anatomically precise text prompts, whose clinical specificity enhances semantic guidance as evidenced by the sharpened heatmap patterns in Fig.~\ref{pic2}. Then, the aligner bridges medical imaging and text domains through four key operations: First, a LoRA-tuned SAM encoder extracts multi-scale visual features $\mathbf{F}_{\text{sam}} \in \mathbb{R}^{B \times C \times H \times W}$, while a CLIP encoder processes Medical-LLM enhanced text prompts into embeddings $\mathbf{F}_{\text{text}} \in \mathbb{R}^{B \times d_{\text{text}}}$, where \( d_{\text{text}} \) represents the dimensionality of the text features and $B$ denoted as the batch size. This process ensures the accurate capture of fine-grained semantic information, providing a solid foundation for subsequent cross-modal alignment. Next, we compute dynamic similarity weights:
\begin{equation}
\mathbf{W}_s = \sigma(\mathcal{P}\left(\frac{\mathbf{F}_{\text{img}}^\top \mathbf{F}_{\text{text}}}{\|\mathbf{F}_{\text{img}}\| \|\mathbf{F}_{\text{text}}\|}\right)),
\end{equation}
 where $\mathcal{P}$ denotes a learnable projection that maps CLIP's global similarity to spatial-wise weights and $\mathbf{W}_s \in \mathbb{R}^{B \times 1 \times L}$. This design explicitly quantifies cross-modal semantic alignment through cosine similarity measurement. Then we construct the spatial attention matrix $\mathbf{A} \in \mathbb{R}^{B \times L \times L}$. It captures inter-pixel relationships, enhanced through layer-normalized dot-product attention:
\begin{equation}
\mathbf{A} = \text{softmax}\left( \frac{{\mathbf{F}_{\text{sam}}^{\text{norm}} \cdot (\mathbf{F}_{\text{sam}}^{\text{norm}}})^\top}{\sqrt{C}} \right),
\end{equation}
where \( C \) represents the number of channels in the feature map.

Finally, the final guidance matrix $\mathbf{G}$ combines attention-refined features with similarity weights through channel-wise broadcasting and dual-level normalization:
\begin{equation}
\mathbf{G} = \Gamma_{\text{spatial}}(\Gamma_{\text{channel}}((\mathbf{F}_{\text{sam}} + \mathbf{A}\mathbf{F}_{\text{sam}}) \odot \mathbf{W}_s)),
\end{equation}
where $\Gamma$ denotes layer normalization operators, and $\odot$ represents element-wise multiplication with broadcasting.

\vspace{-0.25cm}
\subsection{Multi-level Feature Fusion}\label{sec:module2}
\vspace{-0.2cm}
Existing approaches that directly predict masks on low-resolution feature maps often lead to blurry boundaries~\cite{zhang2021refinemaskhighqualityinstancesegmentation}, while simple bilinear upsampling loses crucial high-frequency details~\cite{ronneberger2015}. To address these issues, we propose a multi-level fusion module based on learnable feature reorganization.

We implement cross-scale feature fusion with a pyramid upsampling architecture, as shown in Fig.~\ref{fig:1}, and employ bilinear spatial expansion through a stride-2 transposed convolution. The two-stage upsampling process is formulated as:
\vspace{-0.1cm}
\begin{equation}
\mathbf{F}_{\text{up}}^{(t)} = \sigma\left(\text{LN}\left(\text{DeConv}_{2\times2}(\mathbf{F}_{\text{trans}})\right)\right),\quad t \in \{1,2\},
\end{equation}
where $\mathbf{F}_{\text{trans}}$ is the Transformer's output feature map. This hierarchical design with 2$\times$ resolution increments per stage combined with LN-GELU modules effectively regulates gradient flow, significantly suppressing checkerboard artifacts compared to single-step upsampling~\cite{46191}. The channel compression ratio of 4:2:1 is intentionally adopted to enable the stepwise recovery of high-frequency details, while maintaining computational efficiency throughout the process.

To further enhance spatial awareness, we employ deformable convolution to precisely align the guidance matrix $G \in \mathbb{R}^{C \times H \times W}$ with the upsampled features along the spatial dimensions, thereby effectively capturing complementary cues such as edges and contours that are often missed in coarse features ~\cite{dai2017deformable}. Subsequently, a $1\times1$ convolution is applied to compress the channel dimension, enabling the efficient fusion of the guidance information with the upsampled features, which ultimately achieves integrated cross-modal feature enhancement:

\vspace{-0.2cm}
\begin{equation}
F_{fusion} = \phi\left(F_{up}^{(2)}\right) + \psi\left(\text{Align}(G; \theta)\right),
\end{equation}
where $\phi(\cdot)$ denotes a $1\times1$ convolution used for channel reduction, $\psi(\cdot)$ represents an affine transformation applied to the guidance matrix, and $\theta$ comprises the learnable deformation parameters.

\vspace{-0.2cm}
\subsection{Iterative Mask Optimizer}\label{sec:module3}
\vspace{-0.2cm}

To address coarse edges in initial mask predictions~\cite{zhang2021refinemaskhighqualityinstancesegmentation}, we propose an iterative mask optimizer with two core components:

\noindent\textbf{Instance-Adaptive Kernel Generation.}
To balance general feature extraction with instance-specific adaptation, we design a hypernetwork that generates dynamic convolution parameters. For an instance $i$ with a mask encoding $m_i \in \mathbb{R}^C$, a MLP generates dynamic convolution kernel parameters $\Omega_i$ through:
\vspace{-0.2cm}
\begin{equation}
\Omega_i = \text{MLP}(m_i) \odot \mathcal{W}_{base}, \quad \text{where} \quad \Omega_i \in \mathbb{R}^{C_{in} \times C_{out} \times K \times K}.
\end{equation}
In this equation $\mathcal{W}_{base} \in \mathbb{R}^{C_{in} \times C_{out} \times K \times K}$ represents shared base kernels, and $\odot$ represents the channel-wise Hadamard product~\cite{kim2016hadamard}. 
This design achieves adaptability via two key aspects: (1) the base kernel $\mathcal{W}_{\text{base}}$ extracts universal features across diverse instances; and (2) the dynamic weights $\text{MLP}(m_i)$ encode instance-specific geometric information, modulating channel responses through a gating mechanism to effectively accommodate various object morphologies.

\noindent\textbf{Progressive Residual Refinement.}
We implement mask optimization through iterative residual corrections. The operations of $t$-th iteration:
\vspace{-0.2cm}
\begin{equation}
\Delta M_t = \sigma\left(\text{Conv}_{3\times3}^{\Omega_t}\left([M_t \oplus F_{fusion}]\right)\right),\quad M_{t+1} = \text{Clip}(M_t + \lambda \Delta M_t),
\end{equation}
where $M$ represents mask, $\oplus$ denotes channel concatenation, $\lambda$ is a learnable step-size coefficient, and $\text{Clip}(\cdot)$ constrains the output values to the range $[0,1]$. 

\noindent\textbf{Training Objective.} To train our segmentation model, the overall training objective adopts the combination of cross-entropy and Dice similarity: 
\vspace{-0.2cm}
\begin{equation}
\mathcal{L}_{\text{loss}} = \sum{r \in {l,h}} \Big[ (1-\lambda)\mathcal{L}_{\text{CE}}^r + \lambda\mathcal{L}_{\text{Dice}}^r \Big],
\end{equation}
\noindent where $r \in \{l,h\}$ denotes low/high-resolution paths (56$\times$56 and 224$\times$224) and a hyperparameters $\lambda=0.8$ controls their balance. 

\vspace{-0.3cm}
\section{Experiment} \label{sec:exp}
\vspace{-0.1cm}
\subsection{Experimental Setup}
\vspace{-0.2cm}

\textbf{Dataset.} We evaluate on the MICCAI 2015 Synapse Multi-Organ CT datase~\cite{Landman2015} containing 3,779 contrast-enhanced abdominal CT slices (2,212 training). Following SAMed~\cite{zhang2023customized} and H-SAM~\cite{cheng2024unleashing}, we use 18/12 cases for training/test with 224$\times$224 resolution slices. Evaluation covers eight organs: aorta, gallbladder, spleen, kidneys, liver, pancreas, and stomach.

\noindent\textbf{Implementation details.} We implement training on an RTX 4090 GPU with H-SAM-compatible augmentations. The maximum number of training epochs is set to 300, and the AdamW optimizer is used with $\beta_1$, $\beta_2$, and weight decay set to 0.9, 0.999, and 0.1, respectively. Additionally, we follow the same LoRA configuration as SAMed, where the rank of LoRA is set to 4.

\vspace{-0.6cm}
\subsection{Comparisons with State-of-the-art Methods}
\vspace{-0.3cm}

\begin{table}[t]
\centering
\tiny
\caption{Comparison with state-of-the-art models on Synapse multi-organ CT dataset in both few-shot and fully-supervised settings. Greyed values represent our results, while bold values indicate outperformance of SOTA models. mDice: the Mean Dice coefficient; HD95: the 95th percentile of the Hausdorff Distance.}
\renewcommand{\arraystretch}{1.3}
\resizebox{1\textwidth}{!}{
\begin{tabular}{c|c|cccccccc|cc}
\toprule
& \textbf{Method} & \textbf{Spleen} & \textbf{Kidney(R)} & \textbf{Kidney(L)} & \textbf{Gallbladder} & \textbf{Liver} & \textbf{Stomach} & \textbf{Aorta} & \textbf{Pancreas} & \textbf{mDice}$\uparrow$ & \textbf{HD95}$\downarrow$ \\
\midrule
\multirow{5}{*}{\rotatebox[origin=c]{90}{\centering\textbf{10\%}}} 
& AutoSAM~\cite{hu2023efficiently}    & 68.80 & 77.44 & 76.53 & 24.87 & 88.06 & 52.70 & 75.19 & 34.58 & 55.69 & 31.67 \\
& SAM Adapter~\cite{chen2023sam}       & 72.42 & 68.38 & 66.77 & 22.38 & 89.69 & 53.15 & 66.74 & 26.76 & 58.28 & 54.42 \\
& SAMed~\cite{zhang2023customized}       & 87.32 & 80.10 & 82.75 & 70.24 & 93.37 & 73.62 & 86.99 & 67.64 & 80.26 & 28.89 \\
& H-SAM~\cite{cheng2024unleashing}       & 90.87 & 83.89 & 81.99 & 61.59 & 93.69 & 76.07 & 83.26 & 50.92 & 77.79 & 18.03 \\
& PG-SAM & \cellcolor[gray]{.83}88.43 & \cellcolor[gray]{.83}82.06 & \cellcolor[gray]{.83}\textbf{82.23}$\uparrow$ & \cellcolor[gray]{.83}53.75 & \cellcolor[gray]{.83}92.27 & \cellcolor[gray]{.83}\textbf{78.80}$\uparrow$ & \cellcolor[gray]{.83}82.04 & \cellcolor[gray]{.83}46.43 & \cellcolor[gray]{.83}75.75 & \cellcolor[gray]{.83}\textbf{12.35}$\uparrow$ \\
\midrule
\multirow{10}{*}{\rotatebox[origin=c]{90}{\centering\textbf{Fully - Supervised}}} 
& TransUnet~\cite{chen2021transunet}    & 87.23 & 63.13 & 81.87 & 77.02 & 94.08 & 55.86 & 85.08 & 75.62 & 77.48 & 31.69 \\
& SwinUnet~\cite{cao2022swin}            & 85.47 & 66.53 & 83.28 & 79.61 & 94.29 & 56.58 & 90.66 & 76.60 & 79.13 & 21.55 \\
& TransDeepLab~\cite{azad2022transdeeplab} & 86.04 & 69.16 & 84.08 & 79.88 & 93.53 & 61.19 & 89.00 & 78.40 & 80.16 & 21.25 \\
& DAE-Former~\cite{azad2023dae}          & 88.96 & 72.30 & 86.08 & 80.88 & 94.98 & 65.12 & 91.94 & 79.19 & 82.43 & 17.46 \\
& MERIT~\cite{rahman2024multi}           & 92.01 & 84.85 & 87.79 & 74.40 & 95.26 & 85.38 & 87.71 & 71.81 & 84.90 & 13.22 \\
\cmidrule{2-12}
& AutoSAM~\cite{hu2023efficiently}    & 80.54 & 80.02 & 79.66 & 41.37 & 89.24 & 61.14 & 82.56 & 44.22 & 62.08 & 27.56 \\
& SAM Adapter~\cite{chen2023sam}       & 83.68 & 79.00 & 79.02 & 57.49 & 92.68 & 69.48 & 77.93 & 43.07 & 72.80 & 33.08 \\
& SAMed~\cite{zhang2023customized}       & 87.33 & 80.10 & 82.75 & 70.24 & 93.37 & 73.62 & 86.99 & 67.64 & 80.26 & 28.89 \\
& H-SAM~\cite{cheng2024unleashing}       & 92.34 & 85.99 & 87.71 & 69.65 & 95.20 & 86.27 & 87.53 & 72.53 & 84.65 & 7.29 \\
& PG-SAM & \cellcolor[gray]{.83}\textbf{93.12} & \cellcolor[gray]{.83}84.57 & \cellcolor[gray]{.83}\textbf{87.93} & \cellcolor[gray]{.83}\textbf{73.26} & \cellcolor[gray]{.83}\textbf{95.40} & \cellcolor[gray]{.83}\textbf{86.62} & \cellcolor[gray]{.83}\textbf{87.87} & \cellcolor[gray]{.83}71.49 & \cellcolor[gray]{.83}\textbf{84.79}$\uparrow$ & \cellcolor[gray]{.83}7.61 \\
\bottomrule
\end{tabular}
}
\label{tab:comparison}
\vspace{-0.1cm}
\end{table}

\begin{figure}[t]
    \centering
    \includegraphics[width=\textwidth]{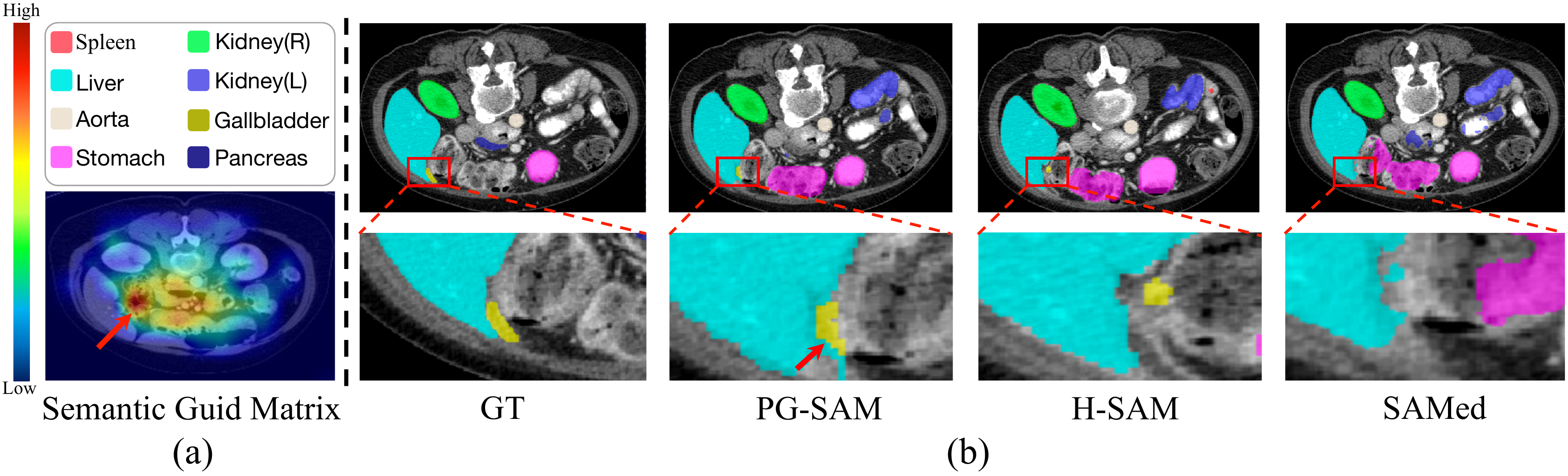}  
    \vspace{-0.5cm}
    \caption{(a) Shows one of the focused areas in our semantic guide matrix; (b) Displays the visualization of segmentation results from various methods on the Synapse dataset especially focus on gallbladder.}
    \label{fig3}
    \vspace{-0.3cm}
\end{figure}

As shown in Table~\ref{tab:comparison}, PG-SAM shows substantial improvements across both few-shot and fully supervised scenarios. Under the 10\% annotation setting, our method exceeds state-of-the-art performance in left kidney segmentation ($\uparrow$0.24\%) and stomach segmentation ($\uparrow$2.73\%), while reducing boundary localization errors to HD95=12.35($\downarrow$5.68\%) compared to the best-performing baseline, demonstrating superior boundary-aware segmentation capabilities. In the fully supervised setting, PG-SAM attains the highest mean Dice coefficient ($\uparrow$0.14\%) among prompt-free SAM variants, with particularly notable improvements in challenging anatomical structures: spleen ($\uparrow$0.78\%), left kidney ($\uparrow$0.22\%), and gallbladder ($\uparrow$3.61\%). Compared to conventional fully supervised methods, PG-SAM achieves dual improvements in both segmentation accuracy and boundary precision: 1) \textbf{Surpasses} TransUNet (+7.31\% Dice) and SwinUNet (+5.66\% Dice) in overall segmentation quality, 2) \textbf{Reduces} boundary errors by HD95$=$7.61 to MERIT ($\downarrow$5.61\%), while maintaining comparable Dice performance ($\downarrow$0.11\%). This demonstrates our method's unique strength in achieving precise boundary delineation without compromising region-wise segmentation accuracy.

\begin{figure}[t]
  \centering
  \begin{minipage}[t]{0.5\textwidth}
    \vspace{0.5pt} 
    \centering
    \includegraphics[width=0.96\textwidth]{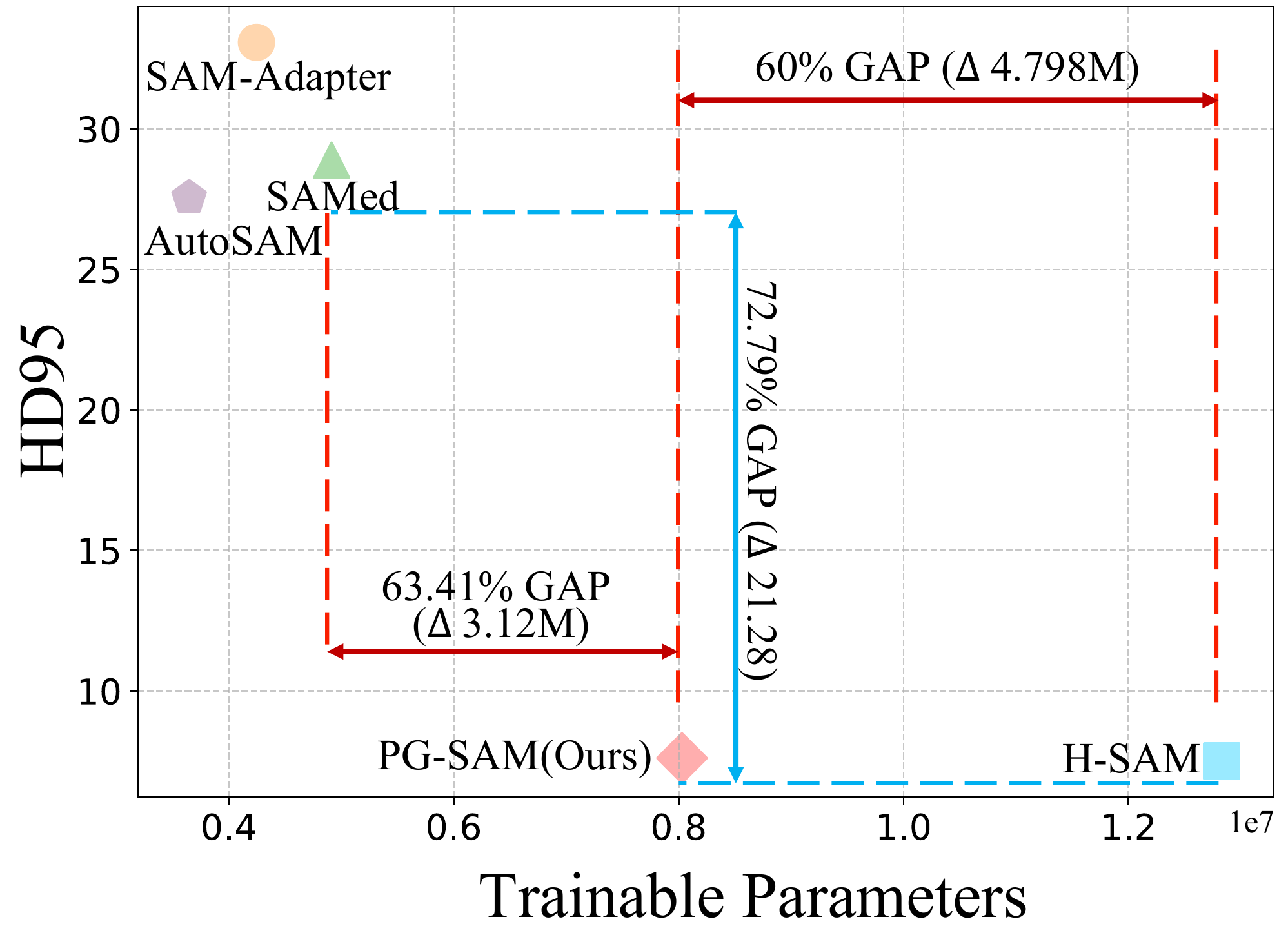}
    \vspace{-0.18cm}
    \caption{Comparing HD95 scores against trainable parameters for different SAMs, with GAP defined as percentage change.}
    \label{fig:trainable}
  \end{minipage}
  \hfill
  \begin{minipage}[t]{0.48\textwidth}
    \centering
    \captionof{table}{Ablation study on the effectiveness of key components in PG-SAM: Fine-Grained Modality Prior Aligner (FGMPA), Multi-level Feature Fusion (MLFF), and Iterative Mask Optimizer (IMO), in terms of mean Dice (mDice) (\%).}
    \resizebox{\textwidth}{!}{
      \begin{tabular}{c|c|c|c|c}
        \toprule
        & \textbf{FGMPA} & \textbf{MLFF} & \textbf{IMO} & \textbf{mDice (\%)} \\
        \midrule
        I   &              &              &              &  72.80\% \\
        II  & $\checkmark$ &              &              &  77.28\%\\
        III & $\checkmark$ & $\checkmark$ &              &  80.10\%\\
        IV  & $\checkmark$ & $\checkmark$ & $\checkmark$ &  84.79\%\\
        \bottomrule
      \end{tabular}
    }
    \label{tab:ablation}
  \end{minipage}
  \vspace{-0.3cm}
\end{figure}

Additionally, PG-SAM achieves remarkable performance efficiency, as evidenced by our experimental results, as demonstrated in Fig.~\ref{fig:trainable}. When comparing Hausdorff Distance 95 (HD95) scores against the number of trainable parameters, PG-SAM demonstrates superior efficiency by maintaining competitive or superior segmentation accuracy while utilizing significantly fewer parameters than existing state-of-the-art approaches.
Despite operating at lower resolution (224$\times$224 vs. 512$\times$512) in few-shot settings which may marginally affect fine detail capture in Medical image, PG-SAM still remains competitive overall.

\vspace{-0.3cm}
\subsection{Qualitative Results}
\vspace{-0.1cm}
In this example, we select the \textbf{gallbladder} as the region of interest for semantic guidance. Fig.~\ref{fig3} (a) shows a heatmap that highlights the key focus areas of the semantic guide matrix, while Fig.~\ref{fig3} (b) presents the corresponding segmentation results. The prior information effectively guides the localization of the gallbladder: SAMed fails to localize it, H-SAM segments it inaccurately, whereas PG-SAM both locates and accurately segments the gallbladder.

\vspace{-0.35cm}
\subsection{Ablation Study}
\vspace{-0.2cm}

To validate the efficacy of the three core components in PG-SAM: FGMPA, MLFF, and IMO, we perform ablation studies on the Synapse dataset, as shown in Table~\ref{tab:ablation}. Starting from Experiment I (Baseline) with a Mean Dice of 72.80\%, adding FGMPA in Experiment II improves cross-modal alignment, leading to a +4.48\% gain. Further integrating MLFF in Experiment III enhances feature fusion, increasing performance by +2.82\%. Finally, incorporating IMO in Experiment IV refines segmentation masks through iterative optimization, achieving the highest Mean Dice of 84.79\%, with a final improvement of +4.69\% over Experiment III. These results demonstrate the synergistic advantage of combining all three components, each contributing to the overall segmentation accuracy.

\vspace{-0.2cm}
\section{Conclusion}
\vspace{-0.2cm}

In this study, we address the limitations of SAM in medical image segmentation, where domain gaps and insufficient textual priors lead to performance degradation. To this end, our proposed PG-SAM integrates medical LLMs to enhance segmentation accuracy. It introduces three key innovations: (1) a fine-grained modality prior aligner for precise anatomical priors, (2) a multi-level feature fusion module that seamlessly integrates global semantic context with local structural details, and (3) an iterative mask optimizer that progressively refines boundary precision. Comprehensive experiments on the Synapse dataset demonstrate that PG-SAM exceeds state-of-the-art performance, enhancing multi-organ segmentation accuracy, especially for complex organs.

\bibliographystyle{splncs04}
\bibliography{main}
\end{document}